# Multilingual LLMs Are Not Multilingual Thinkers: Evidence from Hindi Analogy Evaluation

**Ashray Gupta**[*][†]**, Rohan Joseph**[*][†]**, & Sunny Rai**[◇]

[†]Mahindra University, [◇]University of Pennsylvania
ashray20ucam008@mahindrauniversity.edu.in
rohan18545@mechyd.ac.in

## Abstract

Analogies test a model's ability to infer implicit relationships between concepts, making them a key benchmark for evaluating reasoning capabilities. While large language models (LLMs) are widely evaluated for reasoning in English, their abilities in Indic languages remain understudied, limiting our understanding of whether these models generalize across languages. To address this gap, we introduce a new Hindi Analogy Test Set (HATS), comprising 405 multiple-choice questions sourced from Indian government exams. We benchmark state-of-the-art multilingual LLMs using various prompting strategies and introduce a grounded Chain of Thought approach that leverages cognitive theories of analogical reasoning. This approach improves model performance on Hindi analogy questions. Our experiments show that models perform best with English prompts, irrespective of the prompting strategy. Our test set addresses the lack of a critical resource to evaluate LLM reasoning capabilities in Hindi. The test set is publicly available for research purposes here https://github.com/Inequilazitive/HATS-Hindi_Analogy_Test_Set.

## 1 Introduction

Self-supervised learning enabled language models to learn the notion of *similarity* and *relatedness*. However, abstraction and conceptualization as in analogies, are still a challenge. Growing research on common reasoning tasks including analogies (Ushio et al., 2021; Czinczoll et al., 2022; Bhavya et al., 2022), Winograd Schema Challenge (Liu et al., 2022; Emami et al., 2018), figurative text processing (Joseph et al., 2023; Bogireddy et al., 2023), reflects the trend to teach and evaluate LLMs on these tasks.

Assessing reasoning abilities of LLMs in *low-resource languages* remains challenging (Robinson et al., 2023), primarily due to the scarcity and poor quality of available linguistic data (Khade et al., 2024), as well as the need for improved evaluation methodologies (Valmeekam et al., 2022; Wijesiriwardene et al., 2023; Bender and Koller, 2020). In this paper, we address this resource and knowledge gap by:

- Introducing **HATS**, a test set of 405 of in-situ semantic analogies curated from national and state-level administrative examinations and their preparatory material.

- Benchmarking state-of-the-art multilingual LLMs (see Sec 3.1) with diverse prompting strategies to evaluate LLMs' reasoning abilities in Hindi.

- Proposing a grounded Chain of Thought prompting technique that leverages cognitive theories of analogical reasoning and improves model performance on Hindi analogy tasks (see Sec 3.5.2).

Existing datasets of Hindi analogies are primarily developed by translating English analogies and comprise only syntactic relations (Abdou et al., 2018; Grave et al., 2018). The translated analogies are used to test the quality of Hindi word embeddings (Gaikwad and Haribhakta, 2020) and LLMs trained on Hindi corpus (Kakwani et al., 2020). These datasets lack samples illustrating semantic relations between concepts specific to the Hindi language. This reflects the urgent need for resources to evaluate common reasoning in LLMs in the Indic language.

In this paper, we focus on *proportional analogy* comprising four words of the form $A : B :: C : D$ that is, A is to B as C is to D. Prior works introduced word-family based analogies exploiting

---
[*]These authors contributed equally to this work.

syntactic relations such as *singular-plural* (Abdou et al., 2018). We focus on semantic analogies.

## 2 HATS: Hindi Analogy Test Set

We scraped 405 analogy questions from national and state-level administrative service examinations and preparatory materials, including those for UPSC, SSC, PSC, Clerk, Defense, Railway, and Banking exams, using *BeautifulSoup* (Richardson, 2024). These analogies are designed to assess the aptitude and reasoning abilities of candidates.

> **Example**
>
> भोपाल (Bhopal): मध्य प्रदेश (Madhya Pradesh) :: भुवनेश्वर (Bhubaneshwar): ?
>
> **A** गुजरात (Gujarat)
> **B** उड़ीसा (Odisha)
> **C** राजस्थान (Rajasthan)
> **D** अरुणाचल प्रदेश (Arunachal Pradesh)
>
> **Correct Answer:** उड़ीसा (Odisha), since भुवनेश्वर (Bhubaneshwar) is its capital, just as भोपाल (Bhopal) is the capital of मध्य प्रदेश (Madhya Pradesh).

The original multiple-choice questions appeared in varied formats. We standardized them to the $A : B :: X : Y$ structure and replaced Y with a question mark for model input. We also provide four options that were originally provided with these questions in examinations (See above example).

## 3 Benchmarking LLMs on HATS

### 3.1 Models

We evaluated three state-of-the-art multilingual LLMs: **Aya-expanse-8B** (Dang et al., 2024), **Llama-3.1-8B** (Grattafiori et al., 2024), and **Gemma-2-9B** (Team et al., 2024). These models were selected for their strong performance on multilingual and general-purpose language understanding benchmarks, and their accessibility for academic research (Cohere For AI Team, 2024).

### 3.2 Task A: Find the Most Likely Answer

We create a low-demand (i.e., forced-choice over a fixed set of answer options) task similar to (Hu and Frank, 2024) by presenting the model with an analogy truncated at the last colon ($A : B :: X :$). We select the most likely option as the answer using direct probability measurement. Since we avoid metalinguistic judgment, we chose non-instruct variants of models for this task.

We measured the accuracy of the models using normalized success rates (see Table 1). LLaMA outperforms Aya by 7.46% and Gemma by 6.85%. Overall, model performance in this setting remains suboptimal.

### 3.3 Prompt Design and Evaluation for Generation-Based Tasks

This section outlines the shared design principles and evaluation methodology used across Tasks B and C, both of which involve analogy completion using LLMs. The tasks differ in their prompting strategies but rely on a common structure, a system and user prompt template where we present the task-specific instructions and incomplete analogy with multiple-choice options. For these instruction-centric task settings, we utilize instruction-tuned model variants (see Appendix A for model specifications and prompt details).

**Setting:** To assess the impact of language on reasoning, prompts are evaluated under three configurations: (i) *Hindi-only* (both system and user prompts are in Hindi), (ii) *English-only* (both system and user prompts are in English), and (iii) *Mixed* (English system prompt and Hindi user prompt).

**Evaluation:** To mitigate positional bias in multiple-choice evaluations, we apply a cyclic rotation of the answer options. For a question with $n$ options (typically $n = 4$). we generate $n$ variants, each with the options shifted in position. The model answers all $n$ variants, and the final answer is determined by majority voting across its $n$ responses. A question is marked correct only if the majority-selected answer matches the ground truth; otherwise, it is considered incorrect. Detailed results are discussed in Section 3.6.

### 3.4 Task B: $0-$Shot Prompting

Recent surveys and empirical studies highlight zero-shot prompting as a standard baseline for LLM evaluation, often used to benchmark models before exploring few-shot or fine-tuned settings.(Li, 2023). In the experiments carried out by (Reynolds and McDonell, 2021), the authors show that well-crafted zero-shot prompts can, in fact, surpass the performance of few-shot prompts.

This baseline setting mimics the original exam-style format of the test set. For this task all the instructions were presented in the system prompt.

| Model | Llama 3.1–8B | Aya Expanse–8B | Gemma 2–9B |
|---|---|---|---|
| Accuracy | **46.17** | 42.96 | 43.20 |

Table 1: Accuracy (%) on Task A across all HATS samples. Each score represents the percentage of instances where the model correctly identified the answer option with the highest predicted likelihood.

| Sys + User | Prompting | aya-expanse-8B | Llama-3.1-8B-instruct | gemma-2-9b-it |
|---|---|---|---|---|
| Hi+Hi | 0-Shot | **62.71** | **67.90** | 73.08 |
|  | 0-Shot CoT | **62.71** | 67.40 | 74.81 |
|  | Grounded 0-Shot-CoT | 60.74 | 64.93 | 75.31 |
|  | Grounded FS-CoT | 56.04 | 62.96 | **76.54** |
| En+Hi | 0-Shot CoT | **63.70** | 64.69 | **76.05** |
|  | Grounded 0-Shot-CoT | 61.23 | 64.93 | 75.80 |
|  | Grounded FS-CoT | 59.50 | **65.67** | 75.31 |
| En+En | 0-Shot | **65.67** | 71.85 | 78.77 |
|  | 0-Shot CoT | 65.43 | 66.91 | 78.52 |
|  | Grounded 0-Shot-CoT | 65.43 | **74.56** | **79.75** |
|  | Grounded FS-CoT | 61.72 | 74.07 | 77.28 |
|  | FS Translate-CoT | 62.46 | 72.83 | 77.04 |

Table 2: Accuracy (%) across prompting strategies grouped by language setting. CoT = Chain-of-Thought, FS = Few-Shot. Best scores per setting are bolded. Refer to Section A.2.1 for prompt details. Accuracy is calculated only for valid analogies.

Mixed setting was not evaluated separately, as the prompt content is equivalent to English only in practice.

## 3.5 Task C: Chain of Thought Prompting

Prior work shows that prompting the model to reason step-by-step enhances LLM performance (Brown et al., 2020; Wei et al., 2023; Zhang et al., 2025).

### 3.5.1 0−Shot Chain of Thought

For this task we have taken a similar approach to (Kojima et al., 2023), and appended "Let's think step by step" at the end of the prompt.

### 3.5.2 Grounded 0-Shot Chain of Thought

We build on the (Wang et al., 2023) approach to guide the model's reasoning by presenting a fixed sequence of steps to solve analogies in the prompt. The steps are grounded in cognitive theories of analogical reasoning. Drawing on the (Minnameier, 2010) framework, the prompt integrates abductive structure identification, inductive concept mapping, and adequacy-based evaluation.

### 3.5.3 Grounded Few Shot Chain of Thought

Previous works use few shot examples for prompt based grounding (Mialon et al., 2023). In this task we use the same prompt as in section 3.5.2 with 5 worked out examples. We guided *Claude-3.7-Sonnet*[*] to generate Hindi examples, solved using our Grounded CoT instructions. The examples were verified and corrected by an expert of the Hindi language.

### 3.5.4 Few Shot Chain of Thought (with Translation)

Following the benchmark results, which showed LLMs performed best in English-only settings (see Table 2), we explored whether a translation-based approach could further improve performance on Hindi analogy tasks. Specifically, we implemented a three-step Chain of Thought (CoT) prompting strategy in English (see Sec 3.3):

- **Translation**: Convert the Hindi analogy and options into English.

- **Solution**: Solve the analogy using the method in Section 3.5.2.

- **Mapping**: Identify the correct Hindi option based on the English solution.

---
[*]https://www.anthropic.com/news/claude-3-7-sonnet

We included 5 worked out examples in the prompt. The examples were created using the process described in Section 3.5.3 with updated instructions.

## 3.6 Results

The accuracy scores are presented in Table 2. Prompts in English-only settings consistently led to the highest overall performance. Transitioning from baseline 0−Shot CoT to Grounded 0−Shot CoT resulted in an average improvement of +0.27 points across all models and settings. Gemma was the top performer, achieving the highest accuracy of 79.75% with Grounded 0−Shot Chain-of-Thought prompting (see Sec 3.5.2) in the English-only setting. LLaMA also performed best with Grounded 0−Shot CoT in the English-only setting, reaching an accuracy of 74.56%. In contrast, Aya was the weakest performer, with its highest score being 65.67%, obtained using 0−Shot prompting (see Sec 3.4) in the English-only setting. Some models struggled to follow instructions in Hindi, resulting in better performance with simpler 0−Shot CoT prompts compared to the more complex Grounded CoT setup.

## 4 Discussion

Gemma consistently outperformed other models by an average margin of 11.46 points across all tasks and exhibited minimal performance drop across different prompt settings. All models performed best when both system and user prompts were in English. Chain-of-Thought (CoT) reasoning boosted accuracy, especially in Few-Shot settings.

- While models reliably identified analogical pairs (A : B), they often failed to transfer the relation correctly to (C : D), highlighting limitations in structured reasoning.

- In the translation task, models like *aya–expanse–9b* and *LLaMA–3.1–8B–IT* frequently mistranslated critical terms. For example, the analogy फूल : माला :: ईंट : ? (Flower : Garland :: Brick : ?) was misinterpreted as Flower : Garland :: Eat : ?, confusing ईंट (Brick) with "Eat" due to phonetic similarity. This error was consistent across all 10 sampled failures.

- Models occasionally defaulted to "I don't know" or "None of the above," even when correct options were available.

- See Table A6 for model response languages across different task settings.

## 5 Conclusion

We introduced a test set **HATS** comprising 405 semantic analogies in Hindi. The benchmarking code and prompts for all tasks will be made publicly available. We designed five tasks to evaluate LLMs reasoning abilities in the Hindi language. These tasks assessed the reasoning abilities of LLMs in natural language and the usability of *translation* in creating low-resource language resources. Our experiments reveal the subpar performance of state-of-the-art LLMs when tested on HATS, highlighting the need to evaluate multilingual models on native language resources to better gauge their usability for non-English languages.

## Limitations

In this study, we utilized smaller versions of the model (8B to 9B) due to resource and hardware constraints, and we anticipate models with higher parameters to perform better.

## Ethics Statement

The test set is built from publicly available national level QPs and preparatory material. This ensures that the data is free from (a) anonymity concerns, (b) obscenities and (c) any stereotyping or bias. We have provided a Hindi language resource to evaluate the reasoning abilities of LLMs with the goal to make AI technology accessible to a wider population. We have not performed model training/finetuning and therefore, no significant carbon footprints were generated. We have chosen open source models for this work.

# A Appendix

## A.1 Model Specifications

The model specifications are provided below. We use the pre-trained models.

- Aya Expanse 8B : We set the $max\_new\_tokens = 1200$, $torch\_dtype = torch.float16$, $device\_map = "auto"$, $do\_sample = False$. The model was loaded using the HuggingFace API with the model name 'CohereForAI/aya-expanse-8b'. The model runs in evaluation mode, which disables gradient updates for inference.

- Llama–3.1–8B-Instruct : We set the $max\_new\_tokens = 1200$, $torch\_dtype = torch.float16$, $device\_map = "auto"$, $do\_sample = False$. The model was loaded using the HuggingFace API with the model name 'meta-Llama/Llama-3.1-8B-Instruct'. The model runs in evaluation mode, which disables gradient updates for inference.

- Gemma-2-9b-it : We set the $max\_new\_tokens = 1200$, $torch\_dtype = torch.float16$, $device\_map = "auto"$, $do\_sample = False$. The model was loaded using the HuggingFace API with the model name 'google/gemma-2-9b-it'. The model runs in evaluation mode, which disables gradient updates for inference.

## A.2 Tables

### A.2.1 Prompts

| Prompts for Analogy Tasks |
| --- |
| Task B: 0−Shot Prompting (from Sec 3.4) |
| Models: Gemma-2-9B-it, Llama-3.1-8B-Instruct, Aya-Expanse-8B |
| Hi-Hi Setting |
| **System Prompt:** <br> सादृश्य पूरा कीजिए : <br> आप अपना उत्तर इस प्रकार समाप्त करेंगे: ###अंतिम उत्तर: <आपके द्वारा चुना हुआ विकल्प> |
| **User Prompt:** <br> भोपाल : मध्य प्रदेश :: भुवनेश्वर : ? <br> (A) गुजरात (B) उड़ीसा (C) राजस्थान (D) अरुणाचल प्रदेश |
| En-En Setting |
| **System Prompt:** <br> Complete the analogy: <br> You will end your answer with: ###Final Answer: <Your chosen option> |
| **User Prompt:** <br> भोपाल : मध्य प्रदेश :: भुवनेश्वर : ? <br> (A) गुजरात (B) उड़ीसा (C) राजस्थान (D) अरुणाचल प्रदेश |

Table A1: Prompts for Task B (0− Shot)

| |
| --- |
| Task C: Chain of Thought Prompting (0−shot) (from Sec 3.5.1) |
| Models: Gemma-2-9B-it, Llama-3.1-8B-Instruct, Aya-Expanse-8B |
| Hi-Hi Setting |

| **System Prompt:** |
| --- |
| सादृश्य पूरा कीजिए : |
| आप अपना उत्तर इस प्रकार समाप्त करेंगे: ###अंतिम उत्तर: <आपके द्वारा चुना हुआ विकल्प> |
| **User Prompt:** |
| भोपाल : मध्य प्रदेश :: भुवनेश्वर : ? |
| (A) गुजरात (B) उड़ीसा (C) राजस्थान (D) अरुणाचल प्रदेश |
| आइये कदम दर कदम सोचें |
| En-En Setting |
| **System Prompt:** |
| Complete the analogy: |
| You will end your answer with: ###Final Answer: <Your chosen option> |
| **User Prompt:** |
| भोपाल : मध्य प्रदेश :: भुवनेश्वर : ? |
| (A) गुजरात (B) उड़ीसा (C) राजस्थान (D) अरुणाचल प्रदेश |
| Let's think step by step. |
| Mixed Setting (En + Hi) |
| **System Prompt:** |
| Complete the analogy: |
| You will end your answer with: ###Final Answer: <Your chosen option> |
| **User Prompt:** |
| भोपाल : मध्य प्रदेश :: भुवनेश्वर : ? |
| (A) गुजरात (B) उड़ीसा (C) राजस्थान (D) अरुणाचल प्रदेश |
| आइये कदम दर कदम सोचें |

Table A2: Prompts for Task C (Chain of Thought 0− shot)

Task C: Grounded Zero-Shot Chain of Thought (from Sec 3.5.2)

Models: Gemma-2-9B-it, Llama-3.1-8B-Instruct, Aya-Expanse-8B

Hi-Hi Setting

**System Prompt:**

आप एक समानता (एनालॉजी) संबंधी प्रश्न हल कर रहे हैं। समानता दो चीजों के बीच तुलना होती है, जो किसी न किसी तरह से एक-दूसरे से मेल खाती हैं।

आपका कार्य इस समानता को पूरा करना है, यानी पहले दो शब्दों के बीच के संबंध को समझकर उसी संबंध को तीसरे शब्द पर लागू करना और यह तय करना कि चौथा शब्द क्या होना चाहिए।

समानता हल करने के लिए इन चरणों का पालन करें: सबसे पहले, पहले दो शब्दों (A और B) के बीच के विशिष्ट संबंध को पहचानें। यह समझें कि A का B से क्या संबंध है।

फिर, उसी संबंध को तीसरे शब्द (C) पर लागू करें और देखें कि चौथा शब्द क्या होना चाहिए।

अंत में, दिए गए विकल्पों में से उस विकल्प का चयन करें जो आपके पहचाने गए संबंध के आधार पर समानता को सही तरीके से पूरा करता है।

प्रत्येक चरण में सावधानीपूर्वक सोचें और अंतिम निर्णय लेने से पहले कई संभावित संबंधों पर विचार करें। अपने तर्क को स्पष्ट रूप से प्रस्तुत करें।

अपने अंतिम उत्तर को इस प्रारूप में दें:

### Final Answer: (X) विकल्प

अब निम्नलिखित समानता को इस तीन-चरणीय दृष्टिकोण से हल करें:

**User Prompt:**
सादृश्यता पूरी करें:
भोपाल : मध्य प्रदेश :: भुवनेश्वर : ?
(A) गुजरात (B) उड़ीसा (C) राजस्थान (D) अरुणाचल प्रदेश
पहले बताई गई तीन-चरणीय विधि का पालन करके इस सादृश्य को हल करें।

<span style="color:purple">En-En Setting</span>

**System Prompt:**
You are solving an analogy problem. An analogy is a comparison between two things that are similar in some way. Your task is to complete the analogy by finding the relationship between the first two terms and applying that same relationship to find what the third term relates to. Follow these steps to solve the analogy:

1. First, identify the specific relationship between the first two terms (A and B). Think about how A relates to B.

2. Next, apply this same relationship to the third term (C) to determine what the fourth term should be.

3. Finally, examine each of the given options and select the one that best completes the analogy based on the relationship you identified.

For each step, think carefully and consider multiple possible relationships before deciding. Be explicit in your reasoning. Present your final answer in the format: ###Final Answer: (X) option_text

Now solve the following analogy using this three-step approach:

**User Prompt:**
Complete the following analogy:
भोपाल : मध्य प्रदेश :: भुवनेश्वर : ?
(A) गुजरात (B) उड़ीसा (C) राजस्थान (D) अरुणाचल प्रदेश
by following the three-step method.

<span style="color:purple">Mixed Setting (En + Hi)</span>

**System Prompt:**
You are solving an analogy problem. An analogy is a comparison between two things that are similar in some way. Your task is to complete the analogy by finding the relationship between the first two terms and applying that same relationship to find what the third term relates to. Follow these steps to solve the analogy:

1. First, identify the specific relationship between the first two terms (A and B). Think about how A relates to B.

2. Next, apply this same relationship to the third term (C) to determine what the fourth term should be.

3. Finally, examine each of the given options and select the one that best completes the analogy based on the relationship you identified.

For each step, think carefully and consider multiple possible relationships before deciding. Be explicit in your reasoning. Present your final answer in the format: ###Final Answer: (X) option_text

Now solve the following analogy using this three-step approach:

---

**User Prompt:**

Complete the analogy:

भोपाल : मध्य प्रदेश :: भुवनेश्वर : ?

(A) गुजरात (B) उड़ीसा (C) राजस्थान (D) अरुणाचल प्रदेश

by following the three-step method

---

Table A3: Prompts for Task C (Grounded 0−Shot Chain of Thought)

Task C: Grounded Few Shot Chain of Thought (from Sec 3.5.3)

Models: Gemma-2-9B-it, Llama-3.1-8B-Instruct, Aya-Expanse-8B

Hi-Hi Setting

**System Prompt:**

आप एक समानता (एनालॉजी) संबंधी प्रश्न हल कर रहे हैं। समानता दो चीजों के बीच तुलना होती है, जो किसी न किसी तरह से एक-दूसरे से मेल खाती हैं।

आपका कार्य इस समानता को पूरा करना है, यानी पहले दो शब्दों के बीच के संबंध को समझकर उसी संबंध को तीसरे शब्द पर लागू करना और यह तय करना कि चौथा शब्द क्या होना चाहिए। समानता हल करने के लिए इन चरणों का पालन करें:

1. सबसे पहले, पहले दो शब्दों (A और B) के बीच के विशिष्ट संबंध को पहचानें। यह समझें कि ⬜ का ⬜ से क्या संबंध है।

2. फिर, उसी संबंध को तीसरे शब्द (C) पर लागू करें और देखें कि चौथा शब्द क्या होना चाहिए।

3. अंत में, दिए गए विकल्पों में से उस विकल्प का चयन करें जो आपके पहचाने गए संबंध के आधार पर समानता को सही तरीके से पूरा करता है।

प्रत्येक चरण में सावधानीपूर्वक सोचें और अंतिम निर्णय लेने से पहले कई संभावित संबंधों पर विचार करें। अपने तर्क को स्पष्ट रूप से प्रस्तुत करें। अपने अंतिम उत्तर को इस प्रारूप में दें:

अंतिम उत्तर: (X) विकल्प यहां कुछ उदाहरण दिए गए हैं: उदाहरण 1:

निम्नलिखित समानता को पूरा करें: गंगा : नदी :: हिमालय : ?

(A) पर्वत

(B) देश

(C) महासागर

(D) मैदान

चरण 1: सबसे पहले, मुझे "गंगा" और "नदी" के बीच विशिष्ट संबंध की पहचान करनी है। गंगा एक विशिष्ट नदी है, और "नदी" इसका वर्ग या श्रेणी है। यह एक ऐसा संबंध है जहां पहला शब्द दूसरे शब्द का एक विशिष्ट उदाहरण है।

चरण 2: अब, मुझे इसी संबंध को "हिमालय" पर लागू करना है। यदि हिमालय गंगा की तरह एक विशिष्ट उदाहरण है, तो मुझे इसका वर्ग या श्रेणी ढूंढनी होगी।

चरण 3: अंत में, मुझे प्रत्येक विकल्प की जांच करनी है:

–पर्वत: हिमालय एक विशिष्ट पर्वत श्रृंखला है, और "पर्वत" इसका वर्ग "पर्वत" है। यह गंगा और नदी के समान पैटर्न का अनुसरण करता है।

–देश: हिमालय कोई देश नहीं है; यह एक भौगोलिक विशेषता है

–महासागर: हिमालय का जल निकायों जैसे महासागरों से कोई संबंध नहीं है।

–मैदान: हिमालय मैदान के विपरीत है; यह एक उच्च भूमि है।

अंतिम उत्तर: (A) पर्वत

उदाहरण 2:

निम्नलिखित समानता को पूरा करें: चावल : खेती :: लोहा : ?

(A) धातु

(B) खनन

(C) निर्माण

(D) व्यापार

चरण 1: सबसे पहले, मुझे "चावल" और "खेती" के बीच विशिष्ट संबंध की पहचान करनी है। चावल एक कृषि उत्पाद है जो खेती की प्रक्रिया से प्राप्त होता है। यह एक उत्पाद और उसे प्राप्त करने की प्रक्रिया के बीच का संबंध है।

चरण 2: अब, मुझे इसी संबंध को "लोहा" पर लागू करना है। यदि लोहा चावल की तरह एक उत्पाद है, तो मुझे लोहा प्राप्त करने की प्रक्रिया ढूंढनी होगी।

चरण 3: अंत में, मुझे प्रत्येक विकल्प की जांच करनी है:

- धातु: यह बताता है कि लोहा क्या है (एक धातु), न कि इसे कैसे प्राप्त किया जाता है।

- खनन: खनन वह प्रक्रिया है जिसके द्वारा लोहा पृथ्वी से प्राप्त किया जाता है, जैसे कि खेती वह प्रक्रिया है जिससे चावल प्राप्त होता है। यह वही संबंध बनाए रखता है।

- निर्माण: यह एक ऐसी प्रक्रिया है जो लोहे का उपयोग करती है, न कि लोहा कैसे प्राप्त किया जाता है।

- व्यापार: यह लोहे के वितरण से संबंधित है, न कि इसके उत्पादन से।

अंतिम उत्तर: (B) खनन

उदाहरण 3:

निम्नलिखित समानता को पूरा करें: दिल्ली : भारत :: टोक्यो : ?

(A) चीन

(B) रूस

(C) जापान

(D) कोरिया

चरण 1: सबसे पहले, मुझे "दिल्ली" और "भारत" के बीच विशिष्ट संबंध की पहचान करनी है। दिल्ली भारत की राजधानी है। यह एक राजधानी और उसके देश के बीच का संबंध है। चरण 2: अब, मुझे इसी संबंध को "टोक्यो" पर लागू करना है। यदि टोक्यो दिल्ली की तरह एक राजधानी है, तो मुझे वह देश ढूंढना होगा जिसका टोक्यो राजधानी है। चरण 3: अंत में, मुझे प्रत्येक विकल्प की जांच करनी है:

- चीन: चीन की राजधानी बीजिंग है, टोक्यो नहीं।

- रूस: रूस की राजधानी मॉस्को है, टोक्यो नहीं।

- जापान: टोक्यो जापान की राजधानी है। यह दिल्ली और भारत के समान संबंध बनाए रखता है।

- कोरिया: कोरिया (उत्तर या दक्षिण) की राजधानी प्योंगयांग या सियोल हैं, टोक्यो नहीं।

अंतिम उत्तर: (C) जापान

उदाहरण 4:

निम्नलिखित समानता को पूरा करें: पेंसिल : लिखना :: कैंची : ?

(A) पेपर

(B) काटना

(C) बनाना

(D) सीना

चरण 1: सबसे पहले, मुझे "पेंसिल" और "लिखना" के बीच विशिष्ट संबंध की पहचान करनी है। पेंसिल एक उपकरण है जिसका उपयोग लिखने की क्रिया के लिए किया जाता है। यह एक उपकरण और उसके मुख्य कार्य के बीच का संबंध है।

चरण 2: अब, मुझे इसी संबंध को "कैंची" पर लागू करना है। यदि कैंची पेंसिल की तरह एक उपकरण है, तो मुझे कैंची के मुख्य कार्य को ढूंढना होगा।

चरण 3: अंत में, मुझे प्रत्येक विकल्प की जांच करनी है:

- पेपर: पेपर एक वस्तु है जिस पर काम किया जाता है, न कि एक क्रिया।
- काटना: काटना वह मुख्य क्रिया है जिसके लिए कैंची का उपयोग किया जाता है, जैसे कि लिखना पेंसिल का मुख्य कार्य है।
- बनाना: बनाना कैंची का मुख्य कार्य नहीं है।
- सीना: सीना के लिए आमतौर पर सुई और धागे का उपयोग किया जाता है, न कि कैंची का।

अंतिम उत्तर: (B) काटना

उदाहरण 5:

निम्नलिखित समानता को पूरा करें: शेर : जंगल :: मछली : ?

(A) पिंजरा

(B) समुद्र

(C) रेगिस्तान

(D) खेत

चरण 1: सबसे पहले, मुझे "शेर" और "जंगल" के बीच विशिष्ट संबंध की पहचान करनी है। जंगल वह प्राकृतिक आवास या वातावरण है जहां शेर रहते हैं। यह एक जानवर और उसके प्राकृतिक निवास स्थान के बीच का संबंध है।

चरण 2: अब, मुझे इसी संबंध को "मछली" पर लागू करना है। यदि मछली शेर की तरह एक जानवर है, तो मुझे मछली के प्राकृतिक निवास स्थान या वातावरण को ढूंढना होगा।

चरण 3: अंत में, मुझे प्रत्येक विकल्प की जांच करनी है:

- पिंजरा: पिंजरा एक कृत्रिम वातावरण है जहां जानवरों को रखा जाता है, यह मछली का प्राकृतिक आवास नहीं है।
- समुद्र: समुद्र वह प्राकृतिक जलीय वातावरण है जहां अधिकांश मछलियां रहती हैं, जैसे कि जंगल शेरों का प्राकृतिक आवास है।
- रेगिस्तान: रेगिस्तान एक शुष्क वातावरण है जो मछलियों के लिए उपयुक्त नहीं है।
- खेत: खेत कृषि भूमि है और मछलियों का प्राकृतिक आवास नहीं है।

अंतिम उत्तर: (B) समुद्र

अब निम्नलिखित समानता को इस तीन-चरणीय दृष्टिकोण से हल करें:

**User Prompt:**
सादृश्यता पूरी करें:
भोपाल : मध्य प्रदेश :: भुवनेश्वर : ?
(A) गुजरात (B) उड़ीसा (C) राजस्थान (D) अरुणाचल प्रदेश

En-En Setting

**System Prompt:** You are solving an analogy problem. An analogy is a comparison between two things that are similar in some way. Your task is to complete the analogy by finding the relationship between the first two terms and applying that same relationship to find what the third term relates to.

Follow these steps to solve the analogy:

1. First, identify the specific relationship between the first two terms (A and B). Think about how A relates to B.

2. Next, apply this same relationship to the third term (C) to determine what the fourth term should be.

3. Finally, examine each of the given options and select the one that best completes the analogy based on the relationship you identified.

For each step, think carefully and consider multiple possible relationships before deciding. Be explicit in your reasoning.

Present your final answer in the format: ###Final Answer: (X) option_text

Here are some examples:

Example 1: Complete the analogy: गंगा : नदी :: हिमालय : ?

(A) पर्वत

(B) देश

(C) महासागर

(D) मैदान

Step 1: First, I need to identify the specific relationship between "गंगा" (Ganga) and "नदी" (river). Ganga is a specific river, and "नदी" is its category or classification. This is a relationship where the first term is a specific example of the second term.

Step 2: Next, I need to apply this same relationship to "हिमालय" (Himalaya). If Himalaya is a specific example like Ganga, then I need to find its category or classification.

Step 3: Finally, let me examine each option:

- पर्वत (mountain): Himalaya is a specific mountain range, and "पर्वत" is the category "mountain." This follows the same pattern as Ganga and river.

- देश (country): Himalaya is not a country; it's a geographical feature.

- महासागर (ocean): Himalaya is not related to water bodies like oceans.

- मैदान (plain): Himalaya is the opposite of a plain; it's an elevated landform.

###Final Answer: (A)

Example 2:
Complete the analogy: चावल : खेती :: लोहा : ?

(A) धातु

(B) खनन

(C) निर्माण

(D) व्यापार

Step 1: First, I need to identify the specific relationship between "चावल" (rice) and "खेती" (farming). Rice is an agricultural product that is obtained through the process of farming. This is a relationship between a product and the process used to obtain it.

Step 2: Next, I need to apply this same relationship to "लोहा" (iron). If iron is a product like rice, then I need to find the process used to obtain iron.

Step 3: Finally, let me examine each option:

–धातु (metal): This describes what iron is (a metal), not how it's obtained.

–खनन (mining): Mining is the process by which iron is obtained from the earth, just -as farming is how rice is obtained. This maintains the same relationship.

–निर्माण (construction): This is a process that uses iron, not how iron is obtained.

– व्यापार (trade): This relates to distribution of iron, not its production.

###Final Answer: (B) खनन

Example 3: Complete the analogy: दिल्ली : भारत :: टोक्यो : ?

(A) चीन

(B) रूस

(C) जापान

(D) कोरिया

Step 1: First, I need to identify the specific relationship between "दिल्ली" (Delhi) and "भारत" (India). Delhi is the capital city of India. This is a relationship between a capital city and its country.

Step 2: Next, I need to apply this same relationship to "टोक्यो" (Tokyo). If Tokyo is a capital city like Delhi, then I need to find the country of which Tokyo is the capital.

Step 3: Finally, let me examine each option:

– चीन (China): The capital of China is Beijing, not Tokyo.

– रूस (Russia): The capital of Russia is Moscow, not Tokyo.

– जापान (Japan): Tokyo is the capital of Japan. This maintains the same relationship as Delhi and India.

– कोरिया (Korea): The capitals of Korea (North or South) are Pyongyang or Seoul, not Tokyo.

###Final Answer: (C)

Example 4: Complete the analogy: पेंसिल : लिखना :: कैंची : ?

(A) पेपर

(B) काटना

(C) बनाना

(D) सीना

Step 1: First, I need to identify the specific relationship between "पेंसिल" (pencil) and "लिखना" (writing). A pencil is a tool used for the action of writing. This is a relationship between a tool and its primary function.

Step 2: Next, I need to apply this same relationship to "कैंची" (scissors). If scissors is a tool like a pencil, then I need to find its primary function.

Step 3: Finally, let me examine each option:

- पेपर (paper): Paper is an object that is worked upon, not an action.

- काटना (cutting): Cutting is the primary action for which scissors are used, just as writing is the primary function of a pencil.

- बनाना (making/creating): Making is not the primary function of scissors.

- सीना (sewing): Sewing typically uses needle and thread, not scissors.

###Final Answer: (B) काटना

Example 5:

Complete the analogy: शेर : जंगल :: मछली : ?

(A) पिंजरा

(B) समुद्र

(C) रेगिस्तान

(D) खेत

Step 1: First, I need to identify the specific relationship between "शेर" (lion) and "जंगल" (jungle/forest). A jungle is the natural habitat or environment where lions live. This is a relationship between an animal and its natural habitat.

Step 2: Next, I need to apply this same relationship to "मछली" (fish). If fish is an animal like a lion, then I need to find its natural habitat.

Step 3: Finally, let me examine each option:

- पिंजरा (cage): A cage is an artificial environment where animals are kept, not the natural habitat for fish.

- समुद्र (sea/ocean): The sea/ocean is the natural aquatic environment where most fish live, just as the jungle is the natural habitat for lions.

- रेगिस्तान (desert): A desert is a dry environment that is not suitable for fish.

- खेत (field/farm): A field is agricultural land and not the natural habitat for fish.

###Final Answer: (B) समुद्र

Now solve the following analogy using this three-step approach:

**User Prompt:**
Complete the following analogy:
भोपाल : मध्य प्रदेश :: भुवनेश्वर : ?
(A) गुजरात (B) उड़ीसा (C) राजस्थान (D) अरुणाचल प्रदेश.

Mixed Setting (En + Hi)

**System Prompt:** You are solving an analogy problem. An analogy is a comparison between two things that are similar in some way. Your task is to complete the analogy by finding the relationship between the first two terms and applying that same relationship to find what the third term relates to.

Follow these steps to solve the analogy:

1. First, identify the specific relationship between the first two terms (A and B). Think about how A relates to B.

2. Next, apply this same relationship to the third term (C) to determine what the fourth term should be.

3. Finally, examine each of the given options and select the one that best completes the analogy based on the relationship you identified.

For each step, think carefully and consider multiple possible relationships before deciding. Be explicit in your reasoning.

Present your final answer in the format: ###Final Answer: (X) option_text

Here are some examples:

Example 1: Complete the analogy: गंगा : नदी :: हिमालय : ?

(A) पर्वत

(B) देश

(C) महासागर

(D) मैदान

Step 1: First, I need to identify the specific relationship between "गंगा" (Ganga) and "नदी" (river). Ganga is a specific river, and "नदी" is its category or classification. This is a relationship where the first term is a specific example of the second term.

Step 2: Next, I need to apply this same relationship to "हिमालय" (Himalaya). If Himalaya is a specific example like Ganga, then I need to find its category or classification.

Step 3: Finally, let me examine each option:

- पर्वत (mountain): Himalaya is a specific mountain range, and "पर्वत" is the category "mountain." This follows the same pattern as Ganga and river.

- देश (country): Himalaya is not a country; it's a geographical feature.

- महासागर (ocean): Himalaya is not related to water bodies like oceans.

- मैदान (plain): Himalaya is the opposite of a plain; it's an elevated landform.

###Final Answer: (A)

Example 2:

Complete the analogy: चावल : खेती :: लोहा : ?

(A) धातु

(B) खनन

(C) निर्माण

(D) व्यापार

Step 1: First, I need to identify the specific relationship between "चावल" (rice) and "खेती" (farming). Rice is an agricultural product that is obtained through the process of farming. This is a relationship between a product and the process used to obtain it.

Step 2: Next, I need to apply this same relationship to "लोहा" (iron). If iron is a product like rice, then I need to find the process used to obtain iron.

Step 3: Finally, let me examine each option:

–धातु (metal): This describes what iron is (a metal), not how it's obtained.

–खनन (mining): Mining is the process by which iron is obtained from the earth, just -as farming is how rice is obtained. This maintains the same relationship.

–निर्माण (construction): This is a process that uses iron, not how iron is obtained.

–व्यापार (trade): This relates to distribution of iron, not its production.

###Final Answer: (B) खनन

Example 3: Complete the analogy: दिल्ली : भारत :: टोक्यो : ?

(A) चीन

(B) रूस

(C) जापान

(D) कोरिया

Step 1: First, I need to identify the specific relationship between "दिल्ली" (Delhi) and "भारत" (India). Delhi is the capital city of India. This is a relationship between a capital city and its country.

Step 2: Next, I need to apply this same relationship to "टोक्यो" (Tokyo). If Tokyo is a capital city like Delhi, then I need to find the country of which Tokyo is the capital.

Step 3: Finally, let me examine each option:

–चीन (China): The capital of China is Beijing, not Tokyo.

–रूस (Russia): The capital of Russia is Moscow, not Tokyo.

–जापान (Japan): Tokyo is the capital of Japan. This maintains the same relationship as Delhi and India.

–कोरिया (Korea): The capitals of Korea (North or South) are Pyongyang or Seoul, not Tokyo.

###Final Answer: (C)

Example 4:

Complete the analogy: पेंसिल : लिखना :: कैंची : ?

(A) पेपर

(B) काटना

(C) बनाना

(D) सीना

Step 1: First, I need to identify the specific relationship between "पेंसिल" (pencil) and "लिखना" (writing). A pencil is a tool used for the action of writing. This is a relationship between a tool and its primary function.

Step 2: Next, I need to apply this same relationship to "कैंची" (scissors). If scissors is a tool like a pencil, then I need to find its primary function.

Step 3: Finally, let me examine each option:

- पेपर (paper): Paper is an object that is worked upon, not an action.

- काटना (cutting): Cutting is the primary action for which scissors are used, just as writing is the primary function of a pencil.

- बनाना (making/creating): Making is not the primary function of scissors.

- सीना (sewing): Sewing typically uses needle and thread, not scissors.

###Final Answer: (B) काटना

Example 5:

Complete the analogy: शेर : जंगल :: मछली : ?

(A) पिंजरा

(B) समुद्र

(C) रेगिस्तान

(D) खेत

Step 1: First, I need to identify the specific relationship between "शेर" (lion) and "जंगल" (jungle/forest). A jungle is the natural habitat or environment where lions live. This is a relationship between an animal and its natural habitat.

Step 2: Next, I need to apply this same relationship to "मछली" (fish). If fish is an animal like a lion, then I need to find its natural habitat.

Step 3: Finally, let me examine each option:

- पिंजरा (cage): A cage is an artificial environment where animals are kept, not the natural habitat for fish.

- समुद्र (sea/ocean): The sea/ocean is the natural aquatic environment where most fish live, just as the jungle is the natural habitat for lions.

- रेगिस्तान (desert): A desert is a dry environment that is not suitable for fish.

- खेत (field/farm): A field is agricultural land and not the natural habitat for fish.

###Final Answer: (B) समुद्र

Now solve the following analogy using this three-step approach:

**User Prompt:**
सादृश्यता पूरी करें:
भोपाल : मध्य प्रदेश :: भुवनेश्वर : ?
(A) गुजरात (B) उड़ीसा (C) राजस्थान (D) अरुणाचल प्रदेश

Table A4: Prompts for Task C (Grounded few-Shot Chain of Thought)

Task C: Few Shot Chain of Thought (with Translation) from Sec 3.5.4

Models: Gemma-2-9B-it, Llama-3.1-8B-Instruct, Aya-Expanse-8B

English-only Setting

**System Prompt:** You are solving analogy problems presented in Hindi. An analogy is a comparison between two things that are similar in some way.

Follow these main steps: 1. Translation: Translate the Hindi question and all options to English. 2. Solution: Solve the translated (english) analogy using only English (detailed below). 3. Mapping: Map your English answer back to the correct Hindi option.

For the solution process (step 2), follow these sub-steps: a) Identify the specific relationship between the first two terms (A and B).

b) Apply this same relationship to the third term (C) to determine what the fourth term should be.

c) Examine each of the given options and select the one that best completes the analogy.

IMPORTANT: Use ONLY English words during your solution process (step 2 and its sub-steps).

Only use Hindi when referring to the original question and when giving your final answer.

For each step, think carefully and consider multiple possible relationships. Be explicit in your reasoning.

Present your final answer in the format: ###Final Answer: (X) option_text

Here are some examples:

Example 1:

Complete the analogy: गंगा : नदी :: हिमालय : ?

(A) पर्वत

(B) देश

(C) महासागर

(D) मैदान

Step 1 - Translation:

Question: Ganga : River :: Himalaya : ?

Options:

(A) Mountain

(B) Country

(C) Ocean

(D) Plain

Step 2 - Solution:

a) Relationship identification:

Ganga is a specific river, and River is its category. This is a specific instance to category relationship.

b) Relationship application:

Now I need to apply this relationship to Himalaya. If Ganga is a specific river, then Himalaya would be a specific instance of what category?

c) Option examination:

- Mountain: Himalaya is a specific mountain range, so Mountain is its category. This matches the relationship.

- Country: Himalaya is not a country, it's a geographical feature.

- Ocean: Himalaya is not a body of water, it's a land formation.

- Plain: Himalaya is the opposite of a plain; it's an elevated region.

Step 3 - Mapping:

The answer in English is "Mountain" which corresponds to the Hindi option (A) पर्वत.

###Final Answer: (A)

Example 2:

Complete the analogy: दिल्ली : भारत :: टोक्यो : ?

(A) चीन

(B) रूस

(C) जापान

(D) कोरिया

Step 1 - Translation:

Question: Delhi : India :: Tokyo : ?

Options:

(A) China

(B) Russia

(C) Japan

(D) Korea

Step 2 - Solution:

a) Relationship identification:

Delhi is the capital city of India. This is a capital-country relationship.

b) Relationship application:

Now I need to apply this relationship to Tokyo. I'm looking for the country of which Tokyo is the capital.

c) Option examination:

- China: The capital of China is Beijing, not Tokyo.

- Russia: The capital of Russia is Moscow, not Tokyo.

- Japan: Tokyo is the capital of Japan. This matches the relationship.

- Korea: The capitals of North and South Korea are Pyongyang and Seoul respectively, not Tokyo.

Step 3 - Mapping:

The answer in English is "Japan" which corresponds to the Hindi option (C) जापान.

###Final Answer: (C)

Example 3:

Complete the analogy: चावल : खेती :: लोहा : ?

(A) धातु

(B) खनन

(C) निर्माण

(D) व्यापार

Step 1 - Translation:

Question: Rice : Farming :: Iron : ?

Options:

(A) Metal

(B) Mining

(C) Construction

(D) Trade

Step 2 - Solution:

a) Relationship identification:

Farming is the process by which Rice is produced or obtained. This is a product-production process relationship.

b) Relationship application:

Now I need to apply this relationship to Iron. I'm looking for the process by which Iron is produced or obtained.

c) Option examination:

- Metal: This is a category to which Iron belongs, not a production process.

- Mining: This is the process by which Iron is obtained from the earth, similar to how Farming is used to obtain Rice. This matches the relationship.

- Construction: This is a process that uses Iron, not how it's produced.

- Trade: This relates to the distribution of Iron, not its production.

Step 3 - Mapping:

The answer in English is "Mining" which corresponds to the Hindi option (B) खनन.

###Final Answer: (B)

Example 4:

Complete the analogy: पेंसिल : लिखना :: कैंची : ?

(A) पेपर

(B) काटना

(C) बनाना

(D) सीना

Step 1 - Translation:

Question: Pencil : Writing :: Scissors : ?

Options:

(A) Paper

(B) Cutting

(C) Making

(D) Sewing

Step 2 - Solution:

a) Relationship identification:

A pencil is a tool used for the action of writing. This is a tool-function relationship.

b) Relationship application:

Now I need to apply this relationship to scissors. I'm looking for the primary function of scissors.

c) Option examination:

- Paper: This is an object that is worked upon, not an action.

- Cutting: This is the primary function of scissors, just as writing is the primary function of a pencil.

- Making: This is too general and not the specific function of scissors.

- Sewing: Sewing is done with a needle and thread, not scissors.

Step 3 - Mapping:

The answer in English is "Cutting" which corresponds to the Hindi option (B) काटना.

###Final Answer: (B)

Example 5:

Complete the analogy: शेर : जंगल :: मछली : ?

(A) पिंजरा

(B) समुद्र

(C) रेगिस्तान

(D) खेत

Step 1 - Translation:

Question: Lion : Jungle :: Fish : ?

Options:

(A) Cage

(B) Ocean/Sea

(C) Desert

(D) Field/Farm

Step 2 - Solution:

a) Relationship identification:

A jungle is the natural habitat where lions typically live. This is an animal-habitat relationship.

b) Relationship application:

Now I need to apply this relationship to fish. I'm looking for the natural habitat where fish typically live.

c) Option examination:

- Cage: This is an artificial structure, not a natural habitat.

- Ocean/Sea: This is the natural aquatic environment for most fish, like jungle is for lions.

- Desert: Deserts are dry and unsuitable for fish.

- Field/Farm: This is land used for agriculture, not suitable for fish.

Step 3 - Mapping:

The answer in English is "Ocean/Sea" which corresponds to the Hindi option (B) समुद्र.

###Final Answer: (B)
Now solve the following analogy using the same step-by-step approach. Remember to use ONLY English in your solution process (step 2):

**User Prompt:**
Complete the following analogy:
भोपाल : मध्य प्रदेश :: भुवनेश्वर : ?
(A) गुजरात (B) उड़ीसा (C) राजस्थान (D) अरुणाचल प्रदेश.

Table A5: Prompts for Task C (Few Shot Chain of Thought (with translation))

## A.2.2 Model Response Language across different settings

| Model | Setting (System+User) | 0–Shot | 0–Shot CoT | Grounded 0–Shot CoT | CoT (Few Shot) | CoT (Few Shot–Translate–EN) |
|---|---|---|---|---|---|---|
| aya–expanse–8B | Hi+Hi | Hi | Hi | Hi | Hi | – |
|  | Hi+En | – | Hi | Hi | Hi | – |
|  | En+En | En | En | En | En | En |
| Llama–3.1–8B–instruct | Hi+Hi | Hi | Hi | Hi | Hi | – |
|  | Hi+En | – | Hi | Hi | Hi | – |
|  | En+En | Hi | Hi | En | En | Hi |
| gemma–2–9b–it | Hi+Hi | Hi | Hi | Hi | Hi | – |
|  | Hi+En | – | En | En | En | – |
|  | En+En | En | En | En | En | En |

Table A6: Language in which each model responded across different prompting strategies and language settings